\documentclass[10pt,twocolumn,letterpaper]{article}

\usepackage{wacv}
\usepackage{times}
\usepackage{epsfig}
\usepackage{graphicx}
\usepackage{amsmath}
\usepackage{amssymb}
\usepackage{float}
\usepackage{multicol}
\usepackage[normalem]{ulem}
\usepackage{xcolor,colortbl}
\usepackage{subfigure}
\usepackage{enumitem}

\def\wacvPaperID{679} 

\wacvfinalcopy 

\ifwacvfinal
\def\assignedStartPage{1} 
\fi

\ifwacvfinal
\usepackage[breaklinks=true,bookmarks=false]{hyperref}
\else
\usepackage[pagebackref=true,breaklinks=true,colorlinks,bookmarks=false]{hyperref}
\fi

\ifwacvfinal
\else
\fi

\begin{document}

\title{SLAM in the Field: An Evaluation of Monocular Mapping and Localization on Challenging Dynamic Agricultural Environment}

\author{
Fangwen Shu
\and
Paul Lesur
\and
Yaxu Xie
\and
Alain Pagani
\and
Didier Stricker
\and
DFKI - German Research Center for Artificial Intelligence\\
{\tt\small \{first\_name\}.\{last\_name\}@dfki.de}
}

\maketitle

\begin{abstract}
   This paper demonstrates a system capable of combining a sparse, indirect, monocular visual SLAM, with both offline and real-time Multi-View Stereo (MVS) reconstruction algorithms. This combination overcomes many obstacles encountered by autonomous vehicles or robots employed in agricultural environments, such as overly repetitive patterns, need for very detailed reconstructions, and abrupt movements caused by uneven roads. Furthermore, the use of a monocular SLAM makes our system much easier to integrate with an existing device, as we do not rely on a LiDAR (which is expensive and power consuming), or stereo camera (whose calibration is sensitive to external perturbation e.g. camera being displaced). 
   To the best of our knowledge, this paper presents the first evaluation results for monocular SLAM, and our work further explores unsupervised depth estimation on this specific application scenario by simulating RGB-D SLAM to tackle the scale ambiguity, and shows our approach produces reconstructions that are helpful to various agricultural tasks. Moreover, we highlight that our experiments provide meaningful insight to improve monocular SLAM systems under agricultural settings.
\end{abstract}


\section{Introduction}
\label{sec:intro}

Agricultural robotics \cite{cheein2013agricultural, chen2020advanced, pathan2020artificial, vougioukas2019agricultural} have to function in environments that can be considered adversarial for most SLAM algorithms: abrupt movements, variable illumination, repetitive patterns, and non-rigidness of the environment are all encountered when performing tasks such as harvesting, seeding, agrochemical dispersal, supervision and mapping. Furthermore, while consequent resources have been spent on improving sensor-fusion for SLAM (with e.g. IMU or LiDAR) over the past decades, such systems suffer from sophisticated calibration, added weight, and additional required computational power.
Those points negatively impact the price, power consumption, and algorithm complexity of the robots, all of which are of major importance to manufacturers as well as users.
As such, it is desirable to keep the robot equipped with as few sensors as possible for the given task.

To solve those practical issues, we decided to combine a sparse, feature-based monocular SLAM with both offline and real-time MVS reconstruction algorithm. We show that the SLAM system employed in this work is more reliable for tracking than existing dense SLAM methods, while both reconstruction algorithms' outputs are dense enough for the tasks at hand. We propose the following contributions:

\begin{figure}[t]
  \centering
  \includegraphics[width=0.9\linewidth]{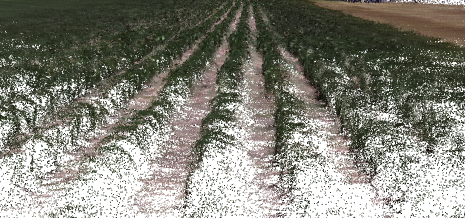}
  \caption{\textbf{The geo-referenced dense point cloud (map)} of soybean field reconstructed from Rosario dataset \cite{pire2019rosario}, sequence 04. 
}
  \label{fig:colmap}
\end{figure}

\begin{itemize}[leftmargin=*, noitemsep]
    \item A usable and efficient dense reconstruction architecture for agricultural mapping and localization with only a single camera.
    \item Exhaustive experiments of indirect visual SLAM systems evaluated on recently released public datasets \cite{chebrolu2017ijrr_dataset, pire2019rosario} aimed at agricultural localization and mapping. Compared to the relative work \cite{comellievaluation, pire2019rosario} which only evaluated stereo setting, we provide the first baseline for monocular SLAM, and improved results for stereo visual SLAM.
    \item  Ablation study of CNN-based self-supervised monocular depth estimation on aforementioned agricultural dataset. The estimated depth is used to simulate RGB-D SLAM with monocular RGB image sequence, which is validated by experimental analysis and competitive results are presented in this work compared to the raw stereo setting.
\end{itemize}

\section{Related work}
\label{sec:related}

\noindent \textbf{Agricultural Robotics} Recent surveys on agricultural robotics \cite{cheein2013agricultural, chen2020advanced, pathan2020artificial, vougioukas2019agricultural} present applications, challenges, and show a growing interest in SLAM system integration. For example, SLAM is a proper solution for occluded GPS (sometimes blocked by dense foliage) \cite{cheein2013agricultural}, crop-relative guidance in open fields, tree-relative guidance in orchards, and more importantly, sensing the crops and its environment \cite{vougioukas2019agricultural}. Various sensors, such as on-board cameras and laser scanners, have been used for extracting features from the crops themselves and use them to localize the robot relative to the crop lines or tree rows in order to auto-steer. 

However, here we focus on related work to monocular vision-based problem instead of discussing the general problem of sensor-fusion, where only a single camera is employed and it is a under-explored problem in agricultural scenario.
\\\\
\noindent \textbf{Dataset} There is a large body of recent and ongoing research regarding SLAM and Visual Odometry for both indoor scenes \cite{Burri25012016, dai2017scannet, handa2014benchmark, sturm2012benchmark, ruiz2017robot} and outdoor urban scenes \cite{carlevaris2016university, geiger2012we, maddern20171, majdik2017zurich}, just to name a few. We do not consider datasets such as \cite{alencastre2018robotics, di2017automatic, haug2014crop, sa2017weednet} as they are extremely specialized for particular tasks like weed/crop classification which are not relevant to this work.
To the best of our knowledge, only two datasets aimed at localization and mapping under agricultural environments are available: the Sugar Beets dataset \cite{chebrolu2017ijrr_dataset} and the Rosario dataset \cite{pire2019rosario}. The former presents a large-scale agricultural robot dataset including downward looking images, captured by a multi-spectral camera and an RGB-D sensor, and we found out it is difficult to track using monocular visual SLAM (downwards looking frames do not cover enough space, and even successive ones have small overlapping regions). The latter consists of 6 sequences recorded in a soybean field, captured by forward looking stereo camera, showing real and challenging cases such as highly repetitive scenes, reflection and burned images caused by direct sunlight and rough terrain, among others. 
\\\\
\noindent \textbf{Advanced Mapping}
As one of the fundamental tasks of mobile robotics, an early work \cite{rovira2008stereo} presented the benefits of building a map of a vehicle's surroundings for precision agriculture. More recently, \cite{dong20174d} has demonstrated a multi-sensor SLAM method for 4D crop monitoring and reconstruction. Another application similar to agricultural mapping is urban mobile mapping system (MMS) presented by \cite{burkhard2012stereovision, cavegn2018robust, cavegn2016image}. There, the choice of dense reconstruction algorithm was more restricted. Commercial software like Pix4D \cite{Pix4D} and Agisoft Metashape \cite{Agisoft} provide sophisticate mapping pipelines with the support of Ground Control Points (GCPs) for indirect geo-referencing and quality control, and \cite{cavegn2016systematic} presented good results of direct geo-referencing by providing camera poses by GPS. However, it is difficult to implement those algorithms on close-range agricultural mapping when image alignment is very challenging due to repetitive texture.
There also exists highly accurate open source methods like COLMAP \cite{schoenberger2016sfm_colmap} and VisualSFM \cite{wu2013towards}, however those frameworks only work offline, and can take up to a few hours to process the data.

When considering real-time dense mapping, it is natural to first try out direct/semi-direct SLAM, in which raw pixels are used for processing, instead of extracting and then matching features using descriptors.
They make it possible to directly reconstruct dense maps as they do not rely on keypoints, but use the entire available data.
This approach has received much attention in the past few years, first with DTAM \cite{dtam}, then with other noteworthy works such as LSD-SLAM \cite{engel14eccv_lsd}, SVO~\cite{Forster2014ICRA_svo} and DSO \cite{engel2016dso}. However, experiments have shown that the aforementioned direct methods have problem initializing at all on the agricultural image sequences (e.g. Rosario dataset).
And the lack of datasets makes the comparison of system performance and robustness in agricultural settings difficult, as we highlight later in this work.

This led us to work with the recently released framework OpenVSLAM \cite{sumikura2019openvslam} which was built upon ORB-SLAM2 \cite{mur2017_orbslam2} without specific change on the core algorithms but provides a new framework with high usability and extensibility. 
After some careful modifications w.r.t agricultural scenarios (which we explain in Section~\ref{sec:implementation}), we are able to initialize and track reliably on challenging agricultural image sequences. Then, initialized by the pose graph generated from SLAM, we adapted COLMAP as an offline dense reconstruction solution, and we ended up working with REMODE \cite{Pizzoli2014ICRA_remode} for real-time dense reconstruction, which was designed as a standalone, monocular reconstruction module running in parallel with another VO (Visual Odometry) module.
\\\\
\noindent \textbf{Simulating RGB-D Sensor} 
Unsupervised learning of depth from unlabelled monocular videos \cite{casser2019depth, godard2017unsupervised, godard2019digging, jiang2020dipe, zhou2017unsupervised} has recently drawn attention as it has notable advantages than the supervised ones and is also the core problem in SLAM \cite{greene2020metrically}. Loosely inspired by the work of CNN-SLAM \cite{tateno2017cnn} and others \cite{li2018undeepvo, loo2019cnn, yang2020d3vo, zhan2018unsupervised}, we integrate Monodepth2 \cite{godard2019digging} as an additional depth predictor to tackle the scale problem of monocular SLAM and the demand of estimating dense depth map during tracking. Such choice is based on the fact that there is no ground truth depth available from dataset Rosario, therefore we have to generate ground truth from stereo image pair using method of SGBM \cite{hirschmuller2007stereo} but only train the depth predictor self-supervised. The predicted depth will be used with monocular image sequence and simulate RGB-D camera which usually has problem working in such outdoor environments.

In this work, we base our experiments on the Rosario dataset \cite{pire2019rosario} as it is appropriate to evaluate monocular SLAM systems. Notice that there is no other relevant work \cite{comellievaluation, pire2019rosario} presents any result of monocular SLAM but only results from stereo SLAM, and our experiment shows reasonably good results on some of the sequence and no tracking lost on all the sequences of Rosario in general. As discussed before, the Sugar Beets dataset \cite{chebrolu2017ijrr_dataset} is discarded due to its irrelevance for dense reconstruction and incompatibility with monocular SLAM.

\section{Implementation Details}
\label{sec:implementation}

First, a monocular feature-based tracking system is used to compute the poses of the camera and acts as the frontend. Then, this information, alongside the original frames, is passed to the backend: a Multi-View Stereo (MVS) reconstruction pipeline that generates a dense point cloud of the agricultural scene, which works either in real-time or offline.
We decided to use OpenVSLAM~\cite{sumikura2019openvslam} which was built upon ORB-SLAM2~\cite{mur2017_orbslam2} as our monocular, feature-based tracker. 
Although literature on SLAM is diverse, most state-of-the-art systems are dense (such as \cite{engel2016dso, Forster2014ICRA_svo}) or fuse more than one type of sensor \cite{Leutenegger2015okvis, Mur_Artal_2017viorb, qin2017vins}, however, ORB-SLAM2 is still as of today the best reference when it comes to feature-based SLAM systems.

\subsection{Monocular, Feature-based Tracker}\label{sec:feature-based}

While the task of reconstructing a dense-map of the environment naturally pushes towards choosing a dense/semi-dense SLAM, experiments have shown that the adversarial nature of agricultural scenes made those systems unreliable.
Meanwhile, we noticed that feature-based methods do not necessarily suffer from the drawbacks inherent to our domain.
Descriptors can be made invariant to lighting and (partially) to blurring, such as \cite{calonder2010brief, Lowe2004sift, rublee2011orb}, which means the tracking is resilient to e.g. holes in the ground, or variable lighting condition due to clouds.
\\\\
\textbf{Auto-Masking of Far Points}
We made modifications to the SLAM system to mask points belonging to the horizon-line dynamically, as they do not bring the depth information necessary to perform tracking.
This is done by estimating the limit between sky (known to be seen at the top of the frame) and the field (known to be at the bottom of the frame), then masking the top of the image until this limit (plus an offset, used to filter all points which close to horizon line), see Figure \ref{fig:remode} (a) and (b) for masking example.
\\\\
\textbf{Monocular Initialization}
The threshold that makes monocular tracking module choose between homography and fundamental matrix model to initialize has been changed so the system picks the fundamental matrix more often:
\begin{equation}
    R_H = \frac{S_H}{S_H + S_F} 
\end{equation}
where $S_H$ and $S_F$ are the scores computed parallel for homography and fundamental matrix, as explained in \cite{mur2015orb}. We found out a robust heuristic to select homography under agricultural settings is $R_H > 0.5$ or even $R_H > 0.8$ in some extreme case. This is purely a domain adaptation change, as we know planar structure are virtually nonexistent in the agricultural scenes we study. The number of ORB features extracted in each frame is also increased drastically to 4000, such as to make the tracking more resilient to potential wrong matches (which arise due to the repetitive nature of the scenes).
\\\\
\textbf{Scale}
Absolute world scale is not observable from a monocular SLAM alone. 
This is a problem we need to tackle as the scale of our reconstruction directly depends on the scale of our tracker.
However, we argue that this problem is easy to solve since it is possible to recover scale information in many ways.
GPS, IMU, or even using some object of known dimensions, can all be used to recover scale information.

We used GPS information in our implementation as it is available in the dataset we worked with. The estimated trajectory from monocular SLAM will be aligned with scale correction as described in \cite{sturm2012benchmark}. Geo-registration of camera center with absolute 3D coordinates will be established before offline MVS reconstruction, as it is described in next section. Moreover, using predicted depth with monocular camera to simulated RGB-D sensor is an alternative to tackle the scale issue, which is evaluated in Section \ref{sec:results}.

\subsection{MVS Dense Reconstruction}

 \begin{figure}[!t]
  \centering
  \includegraphics[width=0.48\textwidth]{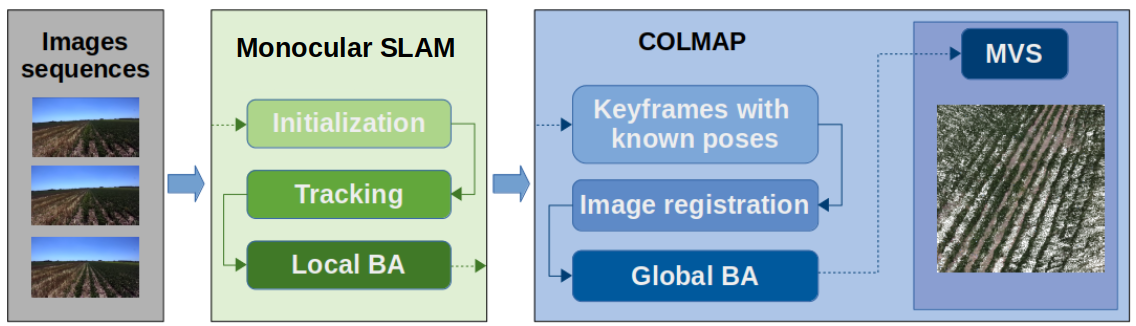}
  \caption{\textbf{The workflow of COLMAP initialized by monocular SLAM for dense reconstruction.} Figure modified from original workflow of \cite{schoenberger2016sfm_colmap}. 
  }
  \label{fig:colmap_wf}
\end{figure}

Once the pose graph has been created, it can be passed to an MVS component that densely reconstructs the environment using the input frames and the corresponding camera poses. This reconstruction can be done either real-time or offline.
Naturally, offline solutions provide much more accurate results, which is of interest for some agricultural applications such as 4D monitoring \cite{dong20174d}. Real-time solutions, on the other hand, provide an initial estimate of the scene which can be used in other tasks where reconstruction does not need to be very dense, such as auto steering.

\begin{figure}[!t]
\centering
\subfigure[Tracked image]{\includegraphics[width=0.2\textwidth]{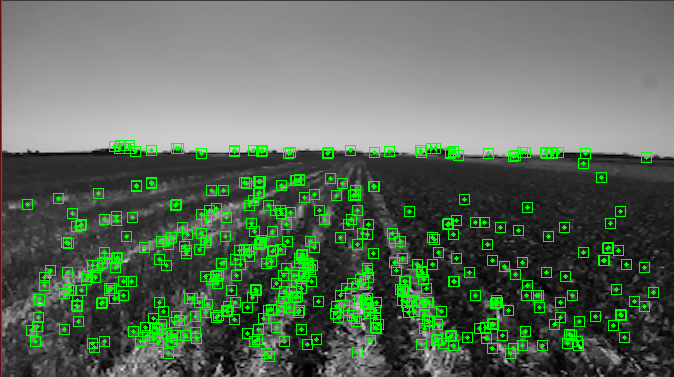}}
\subfigure[Auto-masked image]{\includegraphics[width=0.215\textwidth]{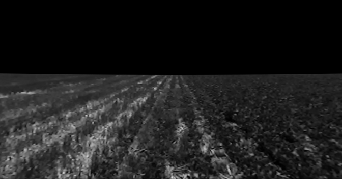}}

\subfigure[Estimated depth map]{\includegraphics[width=0.21\textwidth]{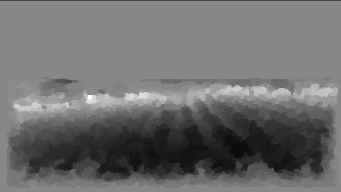}}
\subfigure[Dense point cloud]{\includegraphics[width=0.2\textwidth]{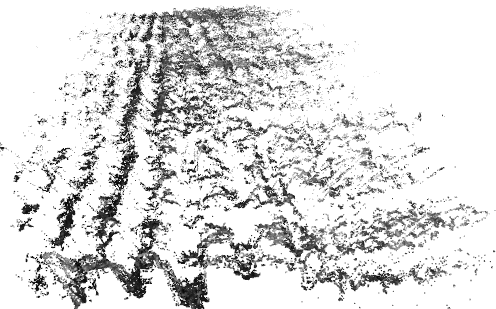}}
\caption{\textbf{Real-time monocular dense reconstruction from REMODE \cite{Pizzoli2014ICRA_remode}} on Rosario \cite{pire2019rosario}, sequence 03, which is running in parallel with monocular SLAM used in this work.}
\label{fig:remode}
\end{figure}

Therefore, the employed SLAM system in this work was embedded with an real-time MVS pipeline, while storing the data for an offline reconstruction once the exploration was finished.
\\\\
\textbf{Offline Dense Reconstruction}
We choose COLMAP~\cite{schoenberger2016mvs_colmap} for the offline, dense reconstruction of our map. It has a well-engineered implementation of Structure-from-Motion (SfM) workflow with Multi-View Stereo (MVS) algorithm \cite{schonberger2016pixelwise}.
The output of SfM is the scene graph includes the camera poses and sparse point cloud, which is considered the same as the output of a sparse SLAM (pose graph). Replacing the SfM part, the standard pipeline is modified by passing the key-frame poses computed by monocular OpenVSLAM to obtain better results. The workflow is illustrated in Figure \ref{fig:colmap_wf}. 
Note that in the image registration stage, we reconstruct sparse point cloud again (not required, but more convenient) to provide neighbourhood information for MVS, in the meantime geo-register image by providing absolute coordinates of camera center. This is similar as the direct geo-referencing using GPS measurement introduced in \cite{cavegn2016systematic}. Thus, the generated point cloud (Figure~\ref{fig:colmap}) is up to real scale and prepared for post-processing, see Figure \ref{fig:geometric_result} for the geometric analysis on the point cloud.
\\\\
\textbf{Online Dense Reconstruction}
There are few real-time MVS reconstruction pipelines, for the obvious reason that accurate dense reconstruction requires a lot of computational power. Still, we are able to integrate REMODE (REgularized MOnocular Depth Estimation) \cite{Pizzoli2014ICRA_remode} with monocular OpenVSLAM to generate maps whose accuracy are high enough for some agricultural tasks such as auto-steer. 
REMODE creates depth filters for every keyframe on per-pixel basis and works on all tracked frames (unlike our offline mode, where only keyframes are used). The filter is initialized with high uncertainty in depth and the mean is set to the average scene depth in the reference frame. Given a set of triangulated noisy depth measurements  $d_1, d_2, ..., d_k$ that correspond to same pixel location, the estimated depth measurement $\tilde{d}_k$  is modeled with a Gaussian + Uniform mixture model distribution \cite{VOGIATZIS2011filter}:
\begin{equation}
    p(\tilde{d}_k|\hat{d}, \rho) = \rho \mathcal{N}(\tilde{d}_k|\hat{d}, \tau_k^2) + (1-\rho) \mathcal{U}(\tilde{d}_k|d_{min}, d_{max})
\end{equation}
where a good depth measurement is assumed to be distributed around the true depth $\hat{d}$ while outlier depth measurements are uniformly distributed within an interval $[d_{min}, d_{max}]$. $\rho$ and $\tau_k^2$ are the probability and the variance of a good measurement. Each new observation is added to its filter, until the covariance is low enough. Thence the filter is considered as having converged, and the 3D point is created in the map using the estimated depth. The output of our system using the online MVS pipeline can be seen Figure~\ref{fig:remode}. 

\subsection{Self-Supervised Monocular Depth Estimation}

We employ Monodepth2 \cite{godard2019digging} as our depth estimation method, which can be self-supervised trained on both monocular videos and stereo pairs. The model is a fully convolutional U-Net \cite{ronneberger2015u} (encoder-decoder structure). When trained on monocular videos, an extra pose estimation network is established to predict the egomotion between image pairs.
\\\\
\textbf{Training}
We trained Monodepth2 with monocular (M), stereo (S) and mixed method (MS) either from pretrained encoder on ImageNet \cite{russakovsky2015imagenet} or starting with high resolution mixed model pretrained (MS*) on KITTI \cite{Geiger2013IJRR} (mono+stereo\_1024x320) which is provided by C. Godard \etal \cite{godard2019digging}. The depth encoders of all models mentioned are ResNet-18 \cite{he2016deep}. All models are trained with a batch size of 6 on single GPU (GEFORCE GTX 1080 Ti) for 10 epochs. The learning rate is set as $1\times10^{-4}$ at the beginning and drops by 0.1 every 4 epochs. Images from the right camera is used in training, while images from left camera are only involved in the computation of loss.
We did not perform horizontal flips as training augmentation, because the principle points of both cameras are not perfectly in the middle of the frame in Rosario dataset. Other data augmentations include random brightness, contrast, saturation, and hue jitter with respective ranges of $\pm0.2$, $\pm0.2$, $\pm0.2$, and $\pm0.1$. 
\\\\
\noindent\textbf{Ground Truth and Metric}
Rosario dataset contains only stereo images, the depth ground truth is generated with SGBM \cite{hirschmuller2007stereo} algorithm and therefore relative noisy. To perform quantitative evaluation, we scaled the predicted depth with the ratio between the median values of predicted depth and ground truth, as done in \cite{godard2019digging}:
\begin{equation}
    D_{predict}^{*} =\frac{median(D_{gt})}{median(D_{predict})} D_{predict}
\end{equation}
where the performance metrics of depth estimation used in this work are: Absolute Relative Error (Abs  Rel), Square  Relative Error (Sq Rel), Root Mean Square Error (RMSE), RMSE log and Accuracy with threshold $(1.25, 1.25^{2}, 1.25^{3})$, marked with red (lower the better) and blue (higher the better) in Table \ref{tab:ablation}. 

In the comparison of the depth estimation accuracy of all six methods in terms of the metrics above, the best results appear both in monocular (M*) and mono+stereo (MS*) training strategies, which results in difficulty of selecting the best model. Thus, we had to simulate RGB-D SLAM with all the possible model at hand (presented in Table \ref{tab:ablation_ATE}). 
The qualitative results of the depth estimation using SGBM and the depth prediction from Monodepth2 with mixed training (MS*) are shown in Figure \ref{fig:cnndisparity}. More results are presented in the supplementary material. The network provides generally accurate depth map, but meets some defects on texture-copy artifacts (e.g. the vehicle windows, 5th row) and on objects with intricate shape (e.g. human bodies and vehicles, 1st row and 4th row).

\section{Experiments and Results}
\label{sec:results}

Three sets of experiments are presented. The first one is the performance benchmarking on dataset Rosario \cite{pire2019rosario} with different SLAM configurations, where we provide the baseline for monocular SLAM and improved results for stereo SLAM, compared to the relative work \cite{comellievaluation, pire2019rosario} which only evaluated stereo setting successfully. Then, along with evaluating monocular SLAM, we establish the ablation study using predicted depth image to simulate RGB-D SLAM. Finally, we discuss the dense point cloud generated in this work.

\subsection{Dataset and Evaluation Methodology}

Exhaustive experiments were established on the Rosario dataset in this work, which is a recently released dataset composed of six different sequences in a soybean field. The available sensor measurements include stereo images (672 $\times$ 376, 15 Hz) and GPS-RTK (5 Hz). The sensors were synchronized and calibrated (both intrinsic and extrinsic). The difficulty of the sequences varies as shown in Table \ref{tab:ATE}. For more details about the agricultural robotic and sensors, please refer to \cite{pire2019rosario}.

The qualitative results of MVS dense reconstruction can be seen in Figure \ref{fig:colmap} and \ref{fig:remode}, where we show the reconstructed point clouds from both offline and online methods. 
Note that the pose graph was geo-registered by providing the absolute position of the camera center before implementing offline MVS on it (typical accuracy of GPS-RTK is around 1cm horizontally and around 2cm vertically). 
For specific tasks like 4D monitoring of crops, this dense point cloud can be used to calculate different kinds of geometric features such as point density.

\begin{figure}[!t]
    \centering
    \includegraphics[width=0.48\textwidth]{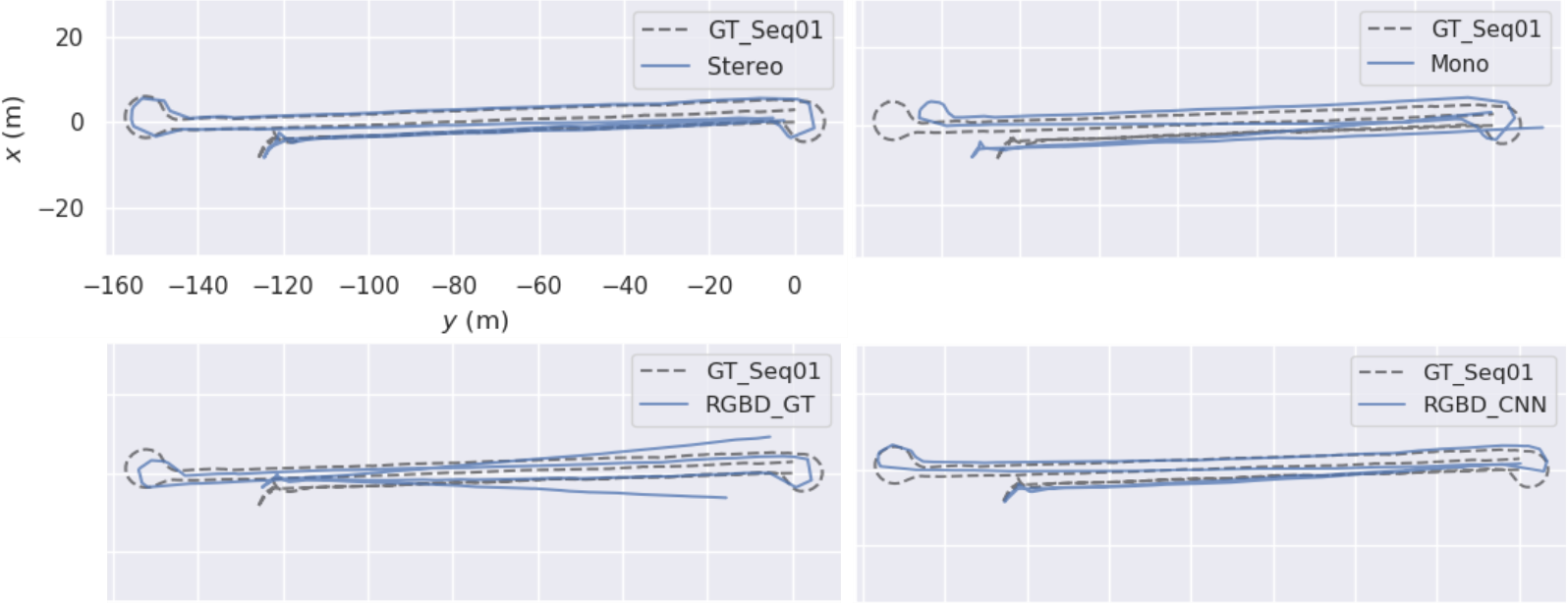}
    \caption{\textbf{Estimated trajectories and the ground truth of Rosario dataset, sequence 01}. The illustrated results are refer to our quantitative results shown in Table \ref{tab:ATE} and Table \ref{tab:ablation_ATE} regarding to OpenVSLAM: Stereo, Mono, Mono+D${}_{GT}^{scaled}$ (RGBD\_GT) and  Mono+D${}_{CNN}^{scaled}$ (RGBD\_CNN, trained model MS*). Results of sequence 02-06 are presented separately in Figure \ref{fig:ATE2}. }
    \label{fig:ATE1}
\end{figure}

\begin{figure*}[!h]
    \centering
    \subfigure[Sequence 02]
    {\includegraphics[width=0.32\textwidth]{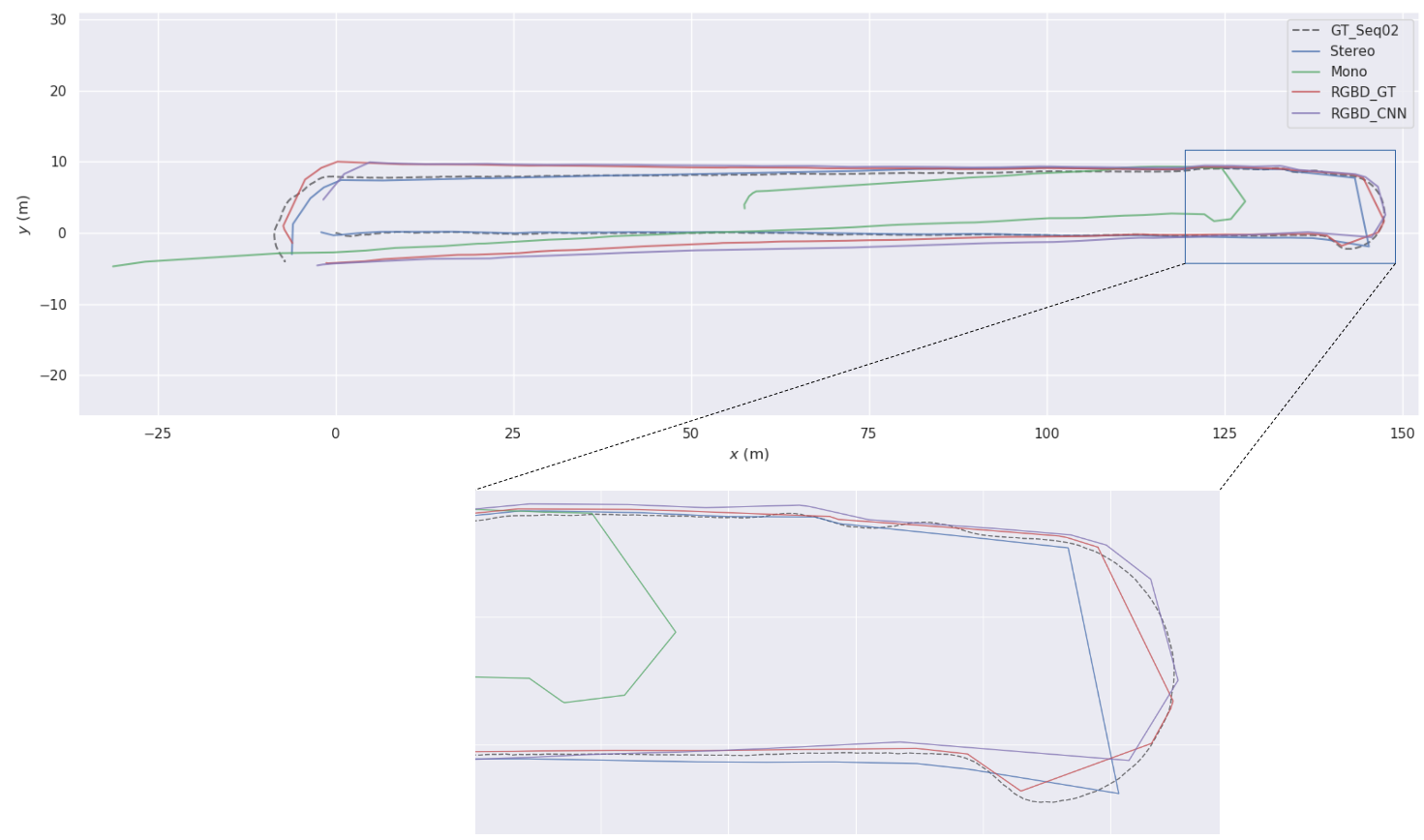}}
    \subfigure[Sequence 03]
    {\includegraphics[width=0.32\textwidth]{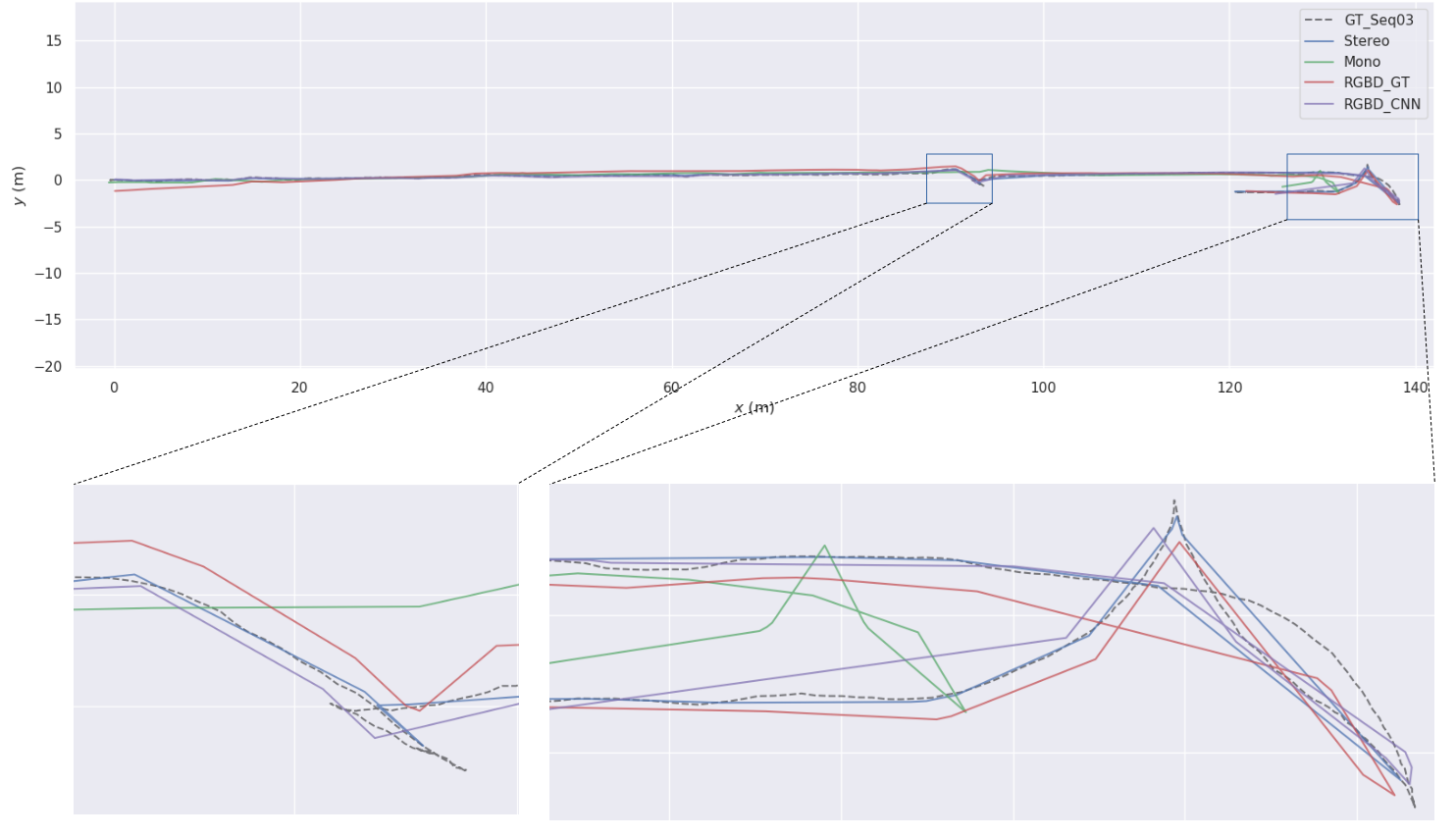}}
    \subfigure[Sequence 04]
    {\includegraphics[width=0.335\textwidth]{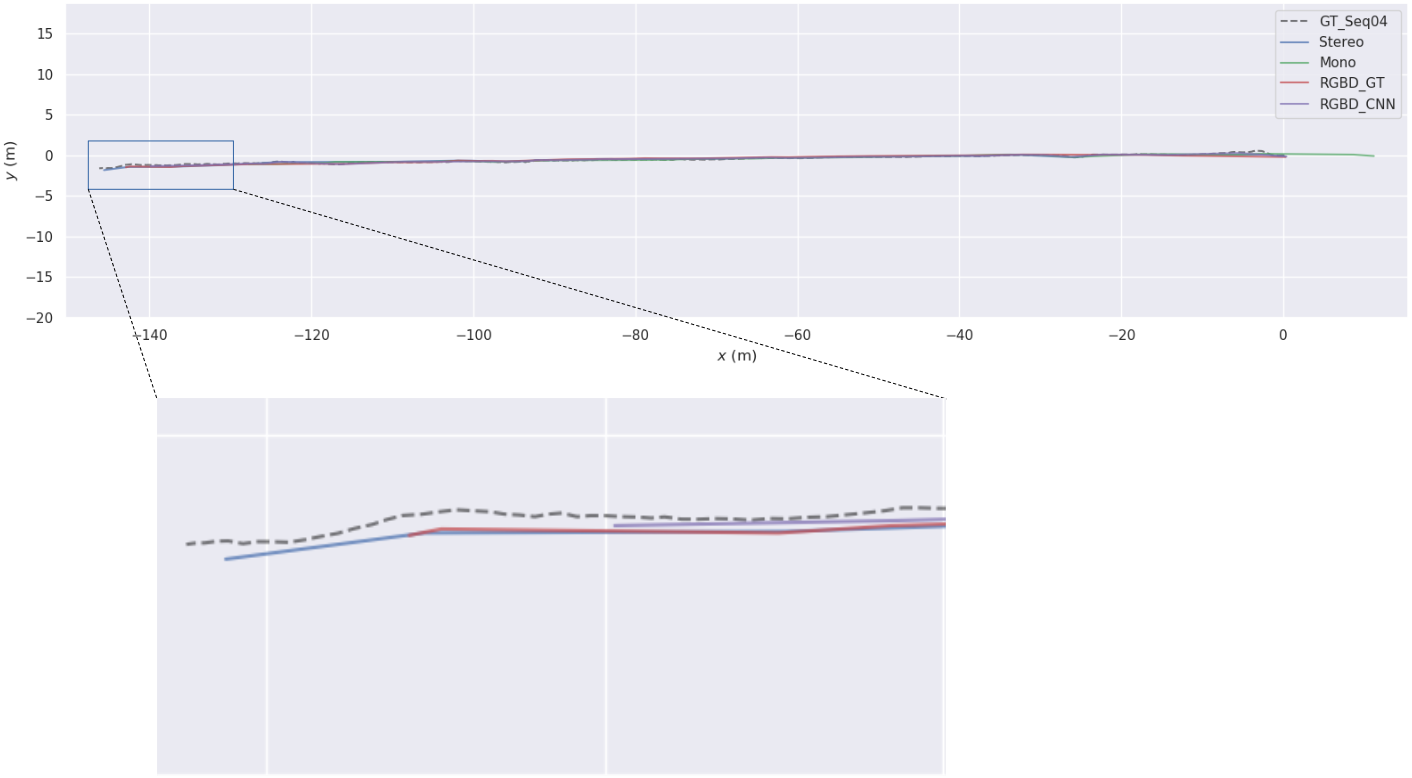}}
    
    \subfigure[Sequence 05]
    {\includegraphics[width=0.33\textwidth]{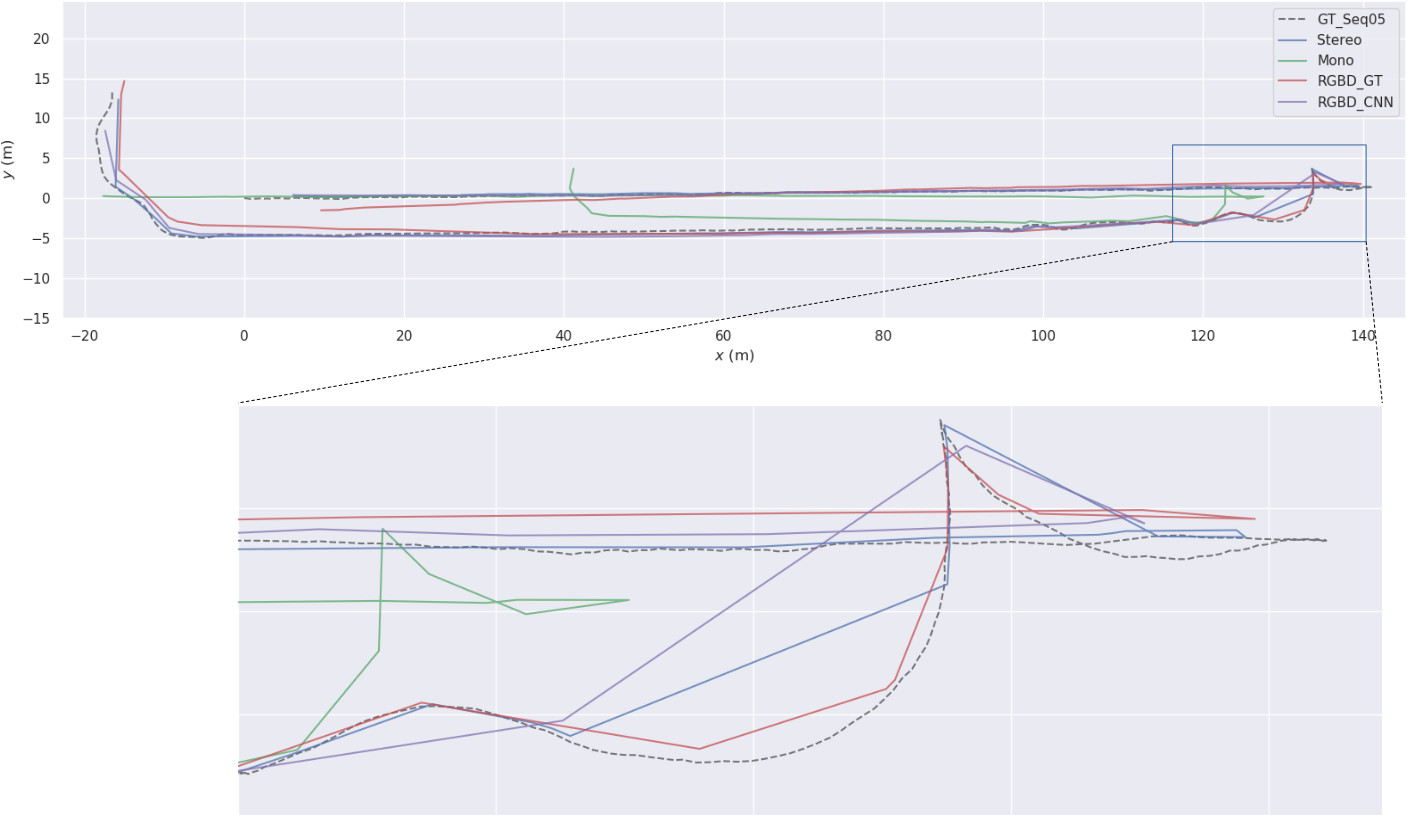}} 
    \subfigure[Sequence 06]
    {\includegraphics[width=0.43\textwidth]{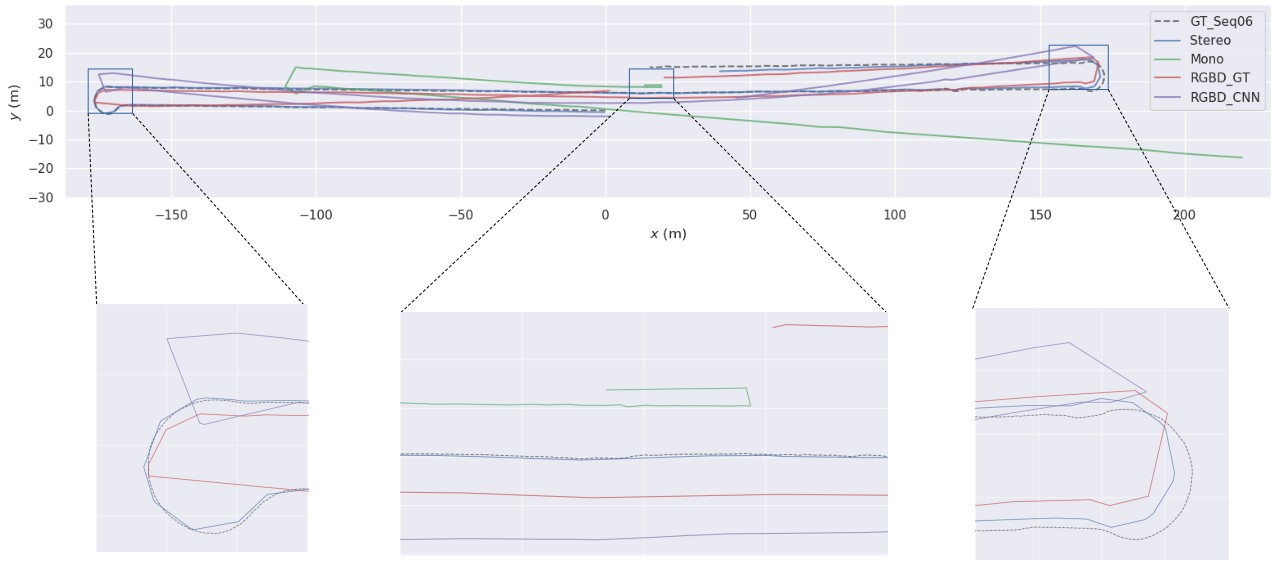}}
    
    \caption{\textbf{Estimated trajectories and the ground truth of Rosario dataset, sequence 02-06}.}
    \label{fig:ATE2}
\end{figure*}

The quantitative results of absolute trajectory error (ATE) estimated from SLAM are shown in Table \ref{tab:ATE} and \ref{tab:ablation_ATE}, corresponding trajectories are illustrated in Figure \ref{fig:ATE1}, \ref{fig:ATE2} and supplementary material. Besides the standard evaluation of ATE for SLAM systems, we also highlight the importance of point density which is used in the field of agricultural mapping (the post-processing results can be seen in Figure \ref{fig:geometric_result}). As introduced in \cite{rovira2008stereo}, a satisfactory methodology to simplify the resolution of 3D field maps while maintaining the key information is through the concept of 3D density and density grids. The idea of the 3D density is rooted in the properties of the conventional density, which establishes a relationship between the mass of a substance and the volume that it occupies:
\begin{equation}
    d = N/V
\end{equation}
Where $N$ indicates the number of points and $V$ indicates the 3D volume with a radius defined by the user. Two practicable approaches to apply the concept of 3D density is to compute either a precise density: the density is estimated by counting for each point the number of neighbors $N$ (inside a sphere of radius $R$); or by computing approximate density: it is then simply estimated by determining the distance to the nearest neighbor (which is generally much faster). This distance is considered as being equivalent to the above spherical neighborhood radius $R$ (and $N$ = 1). In this work, we first compute the precise density, namely, the number of neighbors $N$ with radius of 0.1 m (the absolute scale is known from geo-registration), see Figure \ref{fig:geometric_result}. Thereafter the volume density is calculated simply as: 
\begin{equation}
   d = N/(4/3 \cdot \pi R^3)
\end{equation}

\begin{figure}[!t]
\centering
    \includegraphics[width=.33\linewidth]{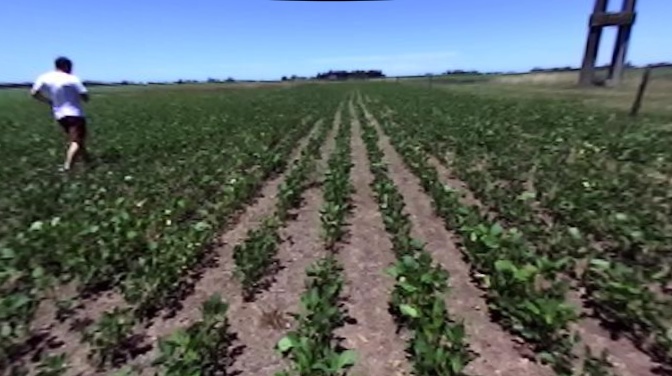}\hfill
    \includegraphics[width=.33\linewidth]{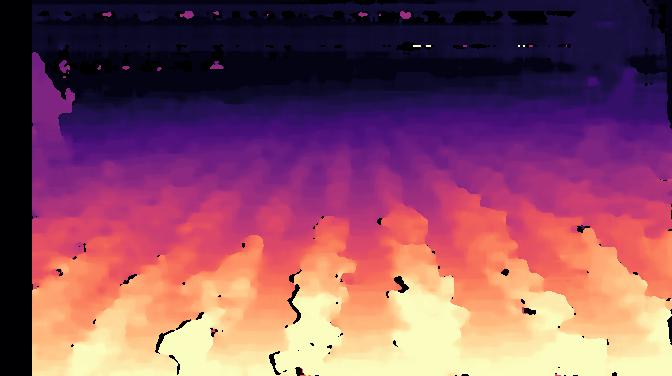}\hfill
    \includegraphics[width=.33\linewidth]{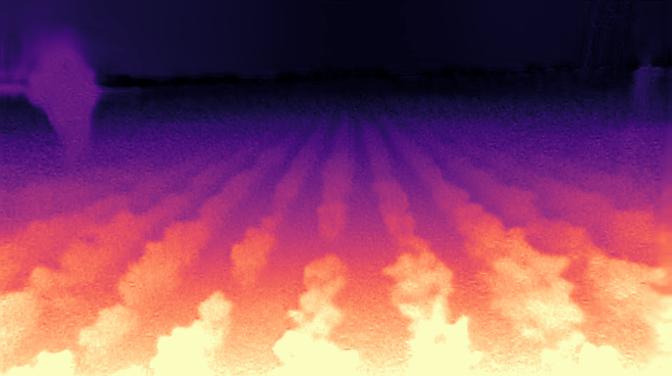}
    
    \includegraphics[width=.33\linewidth]{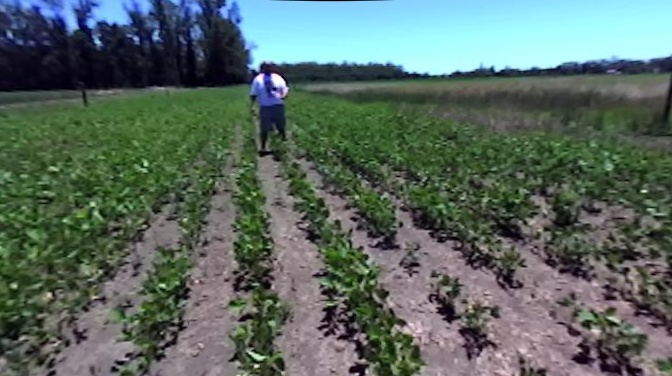}\hfill
    \includegraphics[width=.33\linewidth]{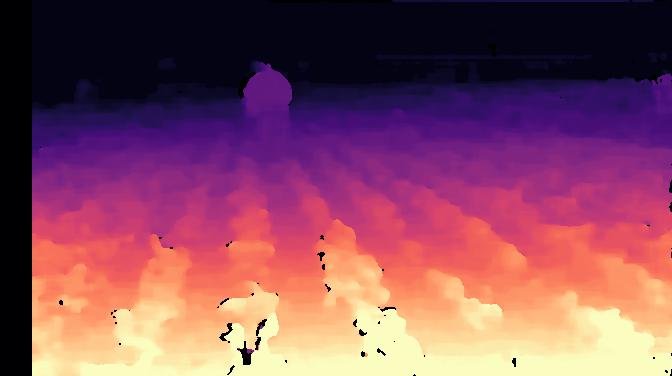}\hfill
    \includegraphics[width=.33\linewidth]{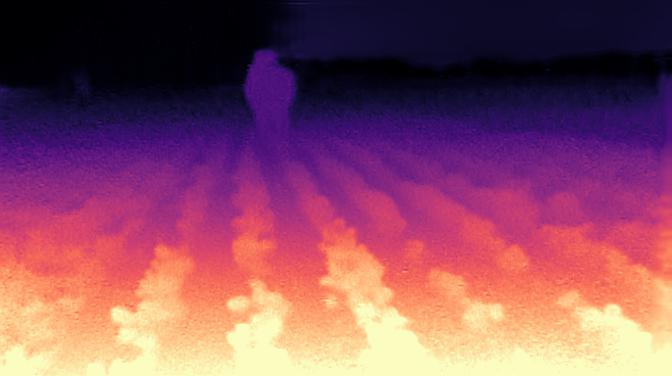}
    
    \includegraphics[width=.33\linewidth]{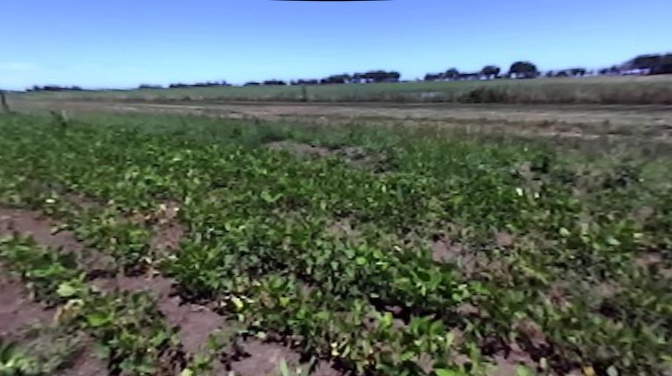}\hfill
    \includegraphics[width=.33\linewidth]{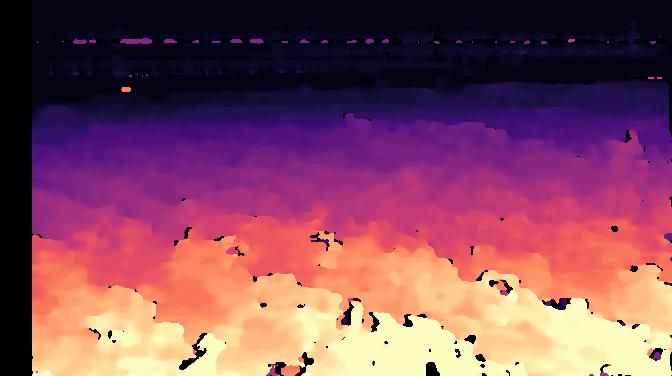}\hfill
    \includegraphics[width=.33\linewidth]{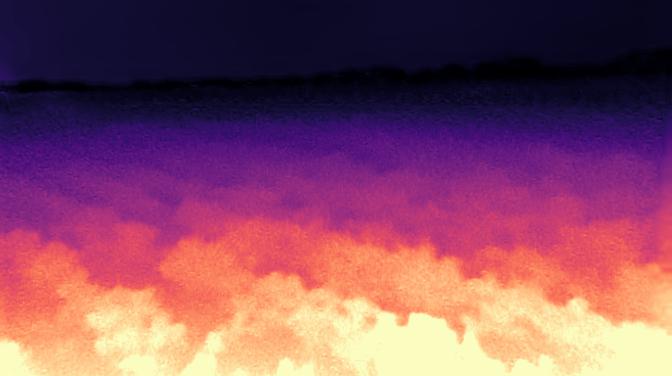}
    
    \includegraphics[width=.33\linewidth]{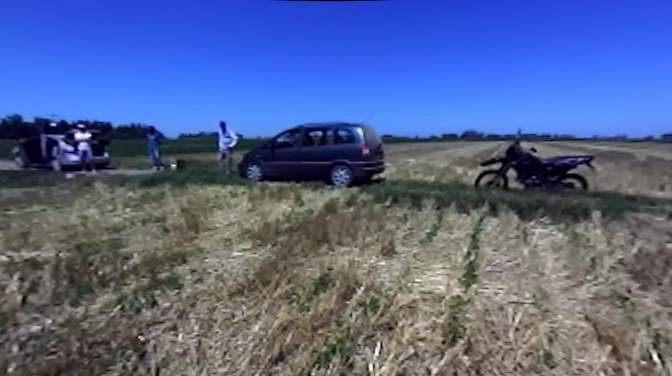}\hfill
    \includegraphics[width=.33\linewidth]{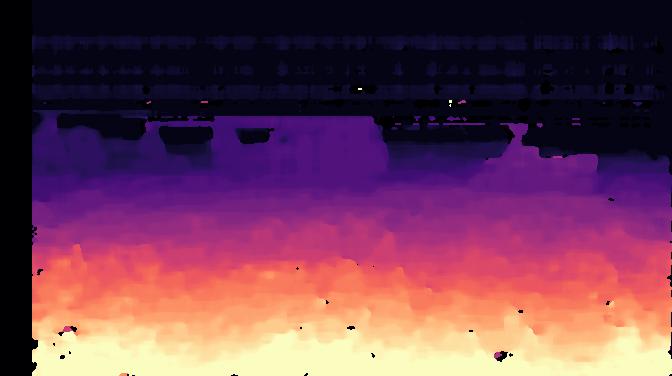}\hfill
    \includegraphics[width=.33\linewidth]{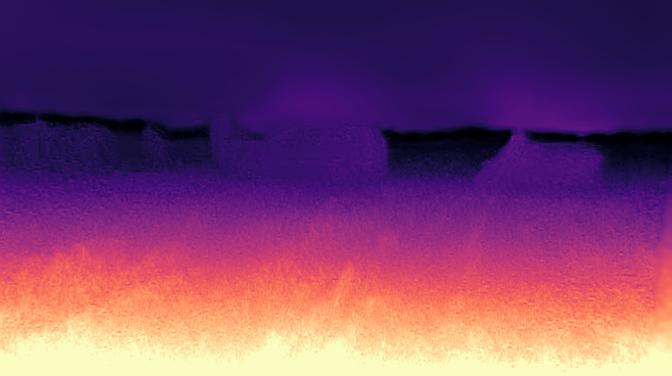}
    
    \includegraphics[width=.33\linewidth]{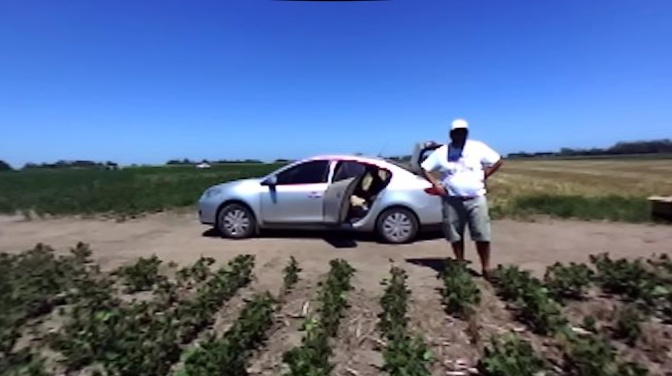}\hfill
    \includegraphics[width=.33\linewidth]{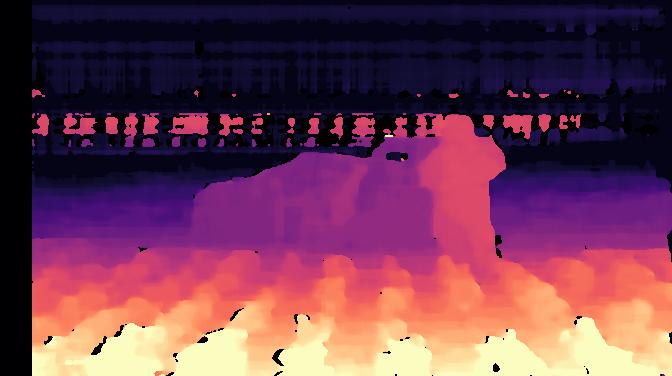}\hfill
    \includegraphics[width=.33\linewidth]{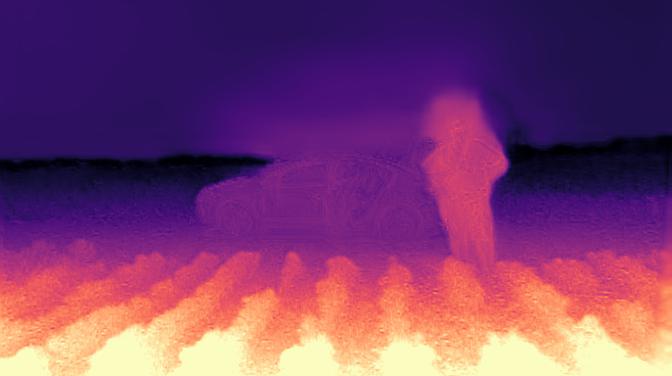}
    
    \includegraphics[width=.33\linewidth]{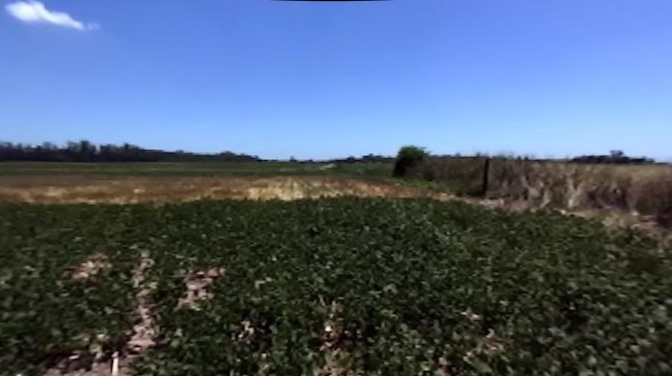}\hfill
    \includegraphics[width=.33\linewidth]{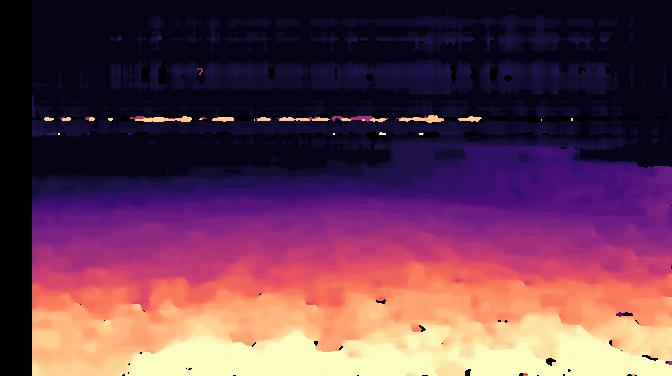}\hfill
    \includegraphics[width=.33\linewidth]{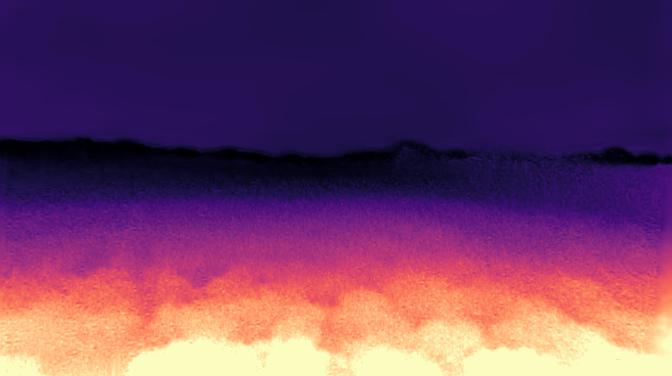}
    
    \includegraphics[width=.33\linewidth]{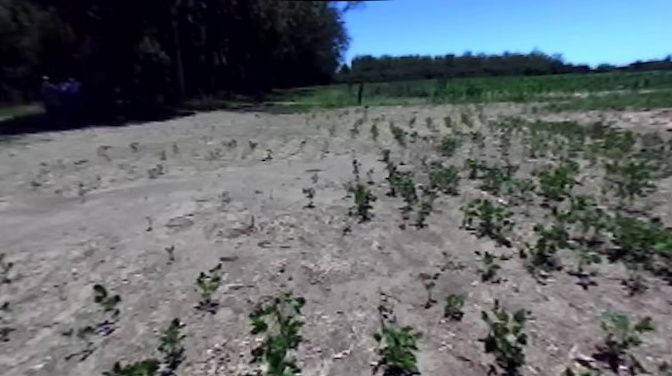}\hfill
    \includegraphics[width=.33\linewidth]{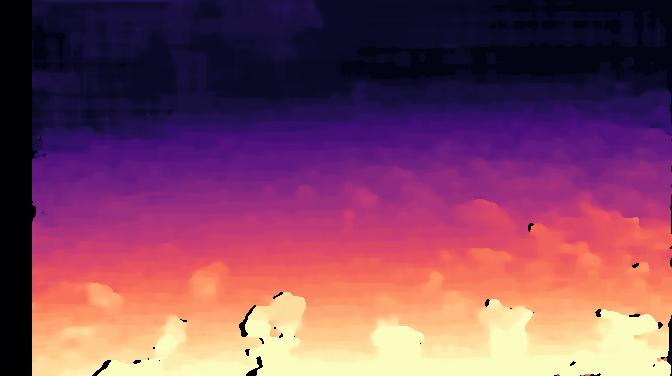}\hfill
    \includegraphics[width=.33\linewidth]{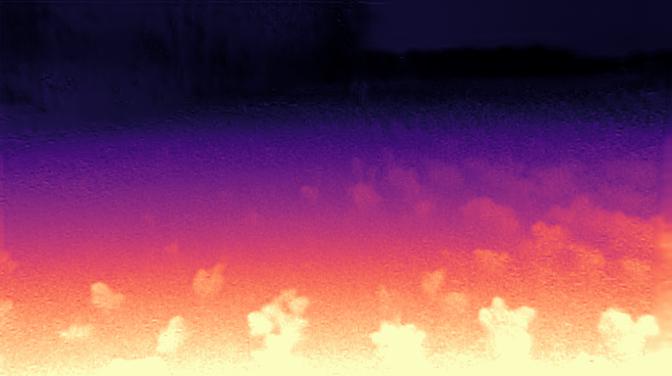}
    \caption{\textbf{Qualitative results of self-supervised monocular depth estimation on Rosario \cite{pire2019rosario}.} First column: selected raw RGB images; Second column: ground truth depth images generated with SGBM \cite{hirschmuller2007stereo}; Third column: predicted depth using Monodepth2 \cite{godard2019digging} with mixed training strategy (MS*). More results please see supplementary material. 
    }
    \label{fig:cnndisparity}
\end{figure}

\begin{table}[!t]
\centering
\small
\scalebox{0.62}{
\begin{tabular}{|l|c||c|c|c|c|c|c|c|}
\hline 
     \multicolumn{2}{|c||}{Dataset Rosario} & S-PTAM \cite{pire2017s} & \multicolumn{2}{c|}{ORB-SLAM2 \cite{mur2017_orbslam2}} &\multicolumn{2}{c|}{OpenVSLAM \cite{sumikura2019openvslam}}  \\
     \hline
     Sequence      &Length  &  Stereo & Stereo & Mono & Stereo & Mono \\
     \hline \hline
     01\_easy      & 615.15 & 3.85 (0.63\%)          &1.41 (0.23\%)  & X & \textbf{1.35 (0.22\%)}   & 10.19 (1.66\%)  \\
     02\_easy      & 320.16 & \textbf{1.80 (0.56\%)} & 2.24 (0.70\%) & X & 1.95 (0.61\%)            & 28.17 (8.80\%)  \\ 
     03\_medium    & 169.45 & 2.37 (1.40\%)          & 3.50 (2.06\%) & X & \textbf{1.75 (1.03\%)}            & 4.29 (2.53\%)   \\ 
     04\_medium    & 152.32 & 1.49 (0.98\%)          & 2.21 (1.45\%) & X & \textbf{1.48 (0.97\%)}            & 6.14 (4.03\%)   \\ 
     05\_difficult & 330.43 & X                      & 2.23 (0.68\%) & X & \textbf{1.65 (0.50\%)}   & 23.66 (7.16\%)  \\
     06\_difficult & 709.42 & X                      & 5.19 (0.73\%) & X & \textbf{3.41 (0.48\%)}   & 91.13 (12.85\%) \\ 
     \hline
\end{tabular}
}
\caption{\textbf{Absolute trajectory error (ATE) [m] (ratio ATE over trajectory length, in \%)} (X stands for tracking failure). 
The results of S-PTAM and Stereo ORB-SLAM2 are extracted from Rosario \cite{pire2019rosario}, Mono ORB-SLAM2 is evaluated in this work with default configuration but the system cannot initialize on any sequence.}
\label{tab:ATE}
\end{table}

\begin{table}[!t]
    \centering
    \small
    \scalebox{0.645}{
    \begin{tabular}{|l||c|c|c|c|c|c|c|}
    \hline
    Method   &\cellcolor{red!25} Abs Rel  &\cellcolor{red!25} Sq Rel   &\cellcolor{red!25} RMSE     &\cellcolor{red!25} RMSE log  &\cellcolor{blue!25}$\sigma<1.25$&\cellcolor{blue!25}$\sigma<1.25^{2}$&\cellcolor{blue!25}$\sigma< 1.25^{3}$   \\ \hline \hline
     M     &   0.151  &   2.110  &   2.961  &   0.205  &   0.913  &   \textbf{0.978}  &   \textbf{0.989}  \\ 
     M*    &   0.150  &   2.118  &   \textbf{2.934}  &   \textbf{0.204}  &   \textbf{0.915}  &   \textbf{0.978}  &   \textbf{0.989}  \\ 
     S     &   0.231  &   1.716  &   5.632  &   0.646  &   0.665  &   0.905  &   0.919  \\
     S*    &   0.235  &   1.737  &   5.650  &   0.644  &   0.657  &   0.900  &   0.919  \\
     M+S   &   \textbf{0.116}  &   0.747  &   3.147  &   0.221  &   0.886  &   0.929  &   0.963  \\
     M+S*  &   \textbf{0.116}  &   \textbf{0.742}  &   3.165  &   0.222  &   0.885  &   0.929  &   0.962  \\ 
     \hline
    \end{tabular}
    }
    \caption{\textbf{Ablation study.} Quantitative results of Monodepth2~\cite{godard2019digging} depth estimation using different variants of training methods on Rosario \cite{pire2019rosario}. \textbf{Legend:} \textbf{S} - Self-supervised stereo supervision; \textbf{M} - Self-supervised  mono supervision; \textbf{*} - start with model pretrained on KITTI \cite{Geiger2013IJRR} (otherwise, the depth encoder is initialized with pretrained weights on ImageNet \cite{russakovsky2015imagenet}). }
    \label{tab:ablation}
\end{table}

\begin{table}[!t]
    \centering
    \small
    \scalebox{0.77}{
    \begin{tabular}{|l|l||c|c|c|c|c|c|}
    \hline
    \multicolumn{2}{|c||}{\textbf{OpenVSLAM}}  & \multicolumn{6}{c|}{ATEs estimated on Dataset Rosario} \\
    \hline
    Setting             & Train   & 01& 02& 03& 04& 05& 06\\
    \hline \hline
    Mono+D${}_{GT}$     & -      &8.32	        &4.94	        &5.70	        &4.31	        &5.92	&13.60 \\
    Mono+D${}_{CNN}$    & M      &X &X &X &X &X &X    \\
    Mono+D${}_{CNN}$    & M*     &10.99	        &12.46	        &16.98	        &14.52	        &13.58	&33.35 \\
    Mono+D${}_{CNN}$    & S      &\textbf{5.21}	&3.80	        &2.79	        &3.01	        &3.26	&8.40  \\
    Mono+D${}_{CNN}$    & S*     &5.25	        &3.78	        &2.73	        &2.96	        &2.90	&8.62  \\
    Mono+D${}_{CNN}$    & MS     &5.44	        &3.57	        &\textbf{2.54}	&\textbf{2.71}	&2.95	&\textbf{7.52}  \\
    Mono+D${}_{CNN}$    & MS*    &5.37	        &\textbf{3.41}	&2.62	        &2.94	        &\textbf{2.63}	&7.79  \\
    \hline \hline
    Mono+D${}_{GT}^{scaled}$     & -      &7.44	        &2.03	      &0.678	        &\textbf{0.25}	&2.39	&\textbf{5.78}\\
    Mono+D${}_{CNN}^{scaled}$    & M      &X &X &X &X &X &X\\
    Mono+D${}_{CNN}^{scaled}$    & M*     &9.30	        &2.91	      &1.04	            &0.72	        &3.60	&10.30\\
    Mono+D${}_{CNN}^{scaled}$    & S      &\textbf{2.31}&1.40	      &0.59	            &0.28	        &2.37	&6.35\\
    Mono+D${}_{CNN}^{scaled}$    & S*     &2.71	        &1.35	      &0.59	            &0.29	        &1.98	&6.95\\
    Mono+D${}_{CNN}^{scaled}$    & MS     &3.61	        &1.35	      &0.60	            &0.26	        &2.24	&6.14\\
    Mono+D${}_{CNN}^{scaled}$    & MS*    &3.29	        &\textbf{1.26}&\textbf{0.57}	&0.26	        &\textbf{1.75}	&6.18\\
    \hline \hline
    Stereo \textbf{(baseline)}   & -      &1.35	&1.95	&1.75	&1.48	&1.65	&3.41\\
    \hline
    \end{tabular}}
    \caption{\textbf{Ablation study.} Quantitative results using estimated depth simulating RGB-D SLAM,  where D${}_{GT}$ and D${}_{CNN}$ indicate whether the depth is generated from stereo image pair as ground truth or estimated from Monodepth2 used in this work, $scaled$ means the estimated trajectory is aligned with scale correction. Baseline (stereo OpenVSLAM) extracted from Table \ref{tab:ATE}.}
    \label{tab:ablation_ATE}
\end{table}

\subsection{Ablation Study}
\label{sec:ablation}

Part of our contribution is evaluating self-supervised depth estimation on agricultural image sequence, along with monocular visual SLAM and simulating RGB-D SLAM. Therefore, a comparison between different training strategy on Rosario \cite{pire2019rosario} is given in Table \ref{tab:ablation} using Monodepth2 \cite{godard2019digging}. We evaluated all the models trained to simulate RGB-D SLAM, where the estimated ATEs are shown in Table \ref{tab:ablation_ATE}. To provide a baseline for future work, there is no specific change in the CNN structure in this work. We discuss the problem regarding to Monodepth2 in Section \ref{sec:eval_depth}.

\subsubsection{Visual SLAM on Rosario}

As shown in Table \ref{tab:ATE}, we present absolute trajectory error (ATE) estimated from monocular OpenVSLAM used in this work, and improved results for stereo SLAM which outperforms the previous baselines from \cite{pire2019rosario} in general. 
Each result of this work was calculated by averaging 5 runs on each sequence. Notice that there is no specific algorithm improvement comparing OpenVSLAM to ORB-SLAM2. Some domain adapted modification on the threshold used in this work was introduced in Section \ref{sec:implementation}. Comparing to the default configuration of monocular SLAM, our modification solved problems of initialization and tracking failure, which is the reason no other work \cite{comellievaluation, pire2019rosario} can present results from monocular SLAM. In fact, sequence 03 and 04 are the two easiest sequences for SLAM as the movement is simple straight forward, where we obtain good results by simulating RGB-D SLAM and competitive good results from Monocular setting. 

\noindent \textbf{Problematic Drifts} Serious drift may occur after the camera inverted its direction (U-turn), such as sequence 02, 05, and 06 evaluated with Mono SLAM (Figure \ref{fig:ATE2}: (a), (d) and (e)). Worst case happen on sequence 06 with Monocular setting, where the scale and estimated trajectory drift dramatically (Figure \ref{fig:ATE2}: (e), the trajectory in green). We also observe that the estimated trajectory on sequence 01, 06 from simulated RGB-D SLAM, drifts after U-turn. We conclude that this is due to the error from ground truth depth generation (see Figure \ref{fig:ATE1} bottom-left, result of RGB\_GT) and self-supervised training (see Figure \ref{fig:ATE1} bottom-right, result of RGB\_CNN).
\\\\
\noindent \textbf{Drifts in Z-Axis}
As we cannot assume a perfect 2D ground plane existing under agricultural scenario and the drifts in z-axis direction have to be considered. The 3D trajectories estimated from SLAM with xyz\_view are illustrated in the supplementary material.
\\\\
\noindent  \textbf{Scale Correction} 
We simulate RGB-D sensor but get better ATE results using scale correction when aligning the trajectory with ground truth (see Table \ref{tab:ablation_ATE}:  Mono+D${}_{GT}^{scale}$ and Mono+D${}_{CNN}^{scale}$), while the results from Mono+D${}_{CNN}^{scale}$ are close to the results from Stereo SLAM, which shows that similar performance can be obtained by simulating RGB-D camera instead of using a raw stereo camera. However, the ground truth depth image should introduce a similar scale as it is generated from the stereo image pair but we still need scale correction, which means the obvious error was introduced during ground truth generation using the method of SGBM \cite{hirschmuller2007stereo}. Comparing all the results of Mono+D${}_{CNN}$, shows that the Monodepth2 also has trouble to learn the accurate scale from agricultural image sequences in a self-supervised fashion, which is further discussed in next Section \ref{sec:eval_depth}.
\\\\
\noindent \textbf{Reproducibility} 
Running Stereo SLAM on Rosario \cite{pire2019rosario} is straightforward and reasonable good results can be obtained, however, we observe that the heuristic threshold used for initializing monocular tracking and the number of ORB features extracted will influence the robustness of the system (as discussed in Section \ref{sec:implementation}). Thus, we provide our experimental results in the supplementary material, where interested readers can find every single value calculated from different SLAM configurations and from 5 test runs on each data sequence.

\subsubsection{Self-Supervised Depth Estimation on Rosario}
\label{sec:eval_depth}

\noindent \textbf{Failure on Textureless Region}
Comparing to urban scenes datasets like KITTI \cite{Geiger2013IJRR}, most frames in Rosario dataset \cite{pire2019rosario} contain a large portion of textureless sky regions. When using stereo training strategy (S/S*), Monodepth2 produces imprecise depth values on low texture regions. When using mixed strategy (MS/MS*), it estimates relative precise depth values on these regions, which is more distinct with foreground objects. Due to the correspondence difficulty, the photometric reconstruction error is ambiguous in large textureless regions. Therefore, a wide range of predicted depth values can produce the same photometric error, which is hard to be optimized based on the left-right consistency assumption \cite{godard2017unsupervised}. 

\begin{figure}[!t]
    \centering
    \subfigure[RGB image]
    {\includegraphics[width=0.24\linewidth]{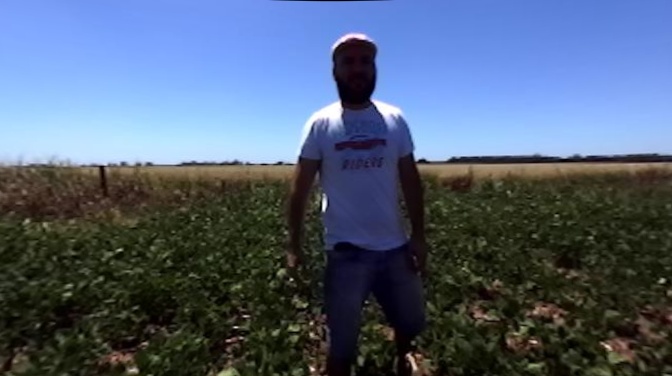}}
    \subfigure[Mono (M*)]
    {\includegraphics[width=0.24\linewidth]{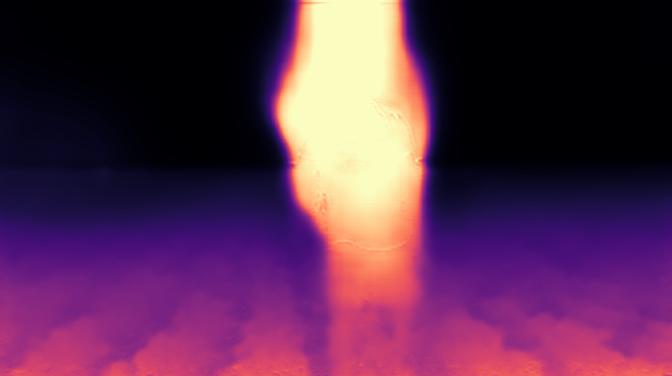}}
    \subfigure[Stereo (S*)]
    {\includegraphics[width=0.24\linewidth]{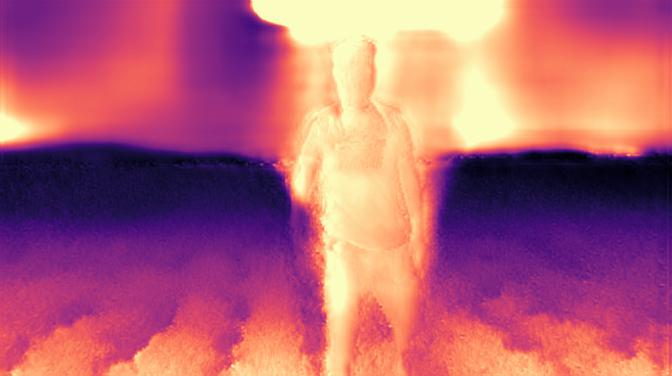}}
    \subfigure[Mixed (MS*)]
    {\includegraphics[width=0.24\linewidth]{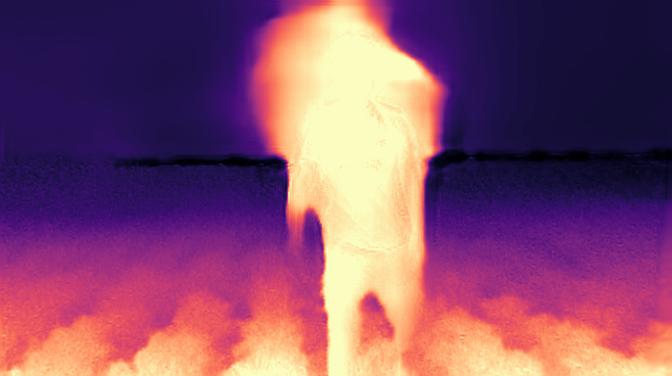}}
    \caption{\textbf{Failure on objects with textureless background.} The network smooths the depth prediction of the sky with the foreground object and results to ambiguous contour of the object.}
    \label{fig:sky}
\end{figure}

The feature-based SLAM system combined with auto-masking of far points (as discussed in Section \ref{sec:feature-based}), tracks no feature point on the textureless region, thus minimizing the negative effects of the unreliable depth estimation. However, we observe some failure cases, which may influence the performance of the SLAM system. As illustrated in Figure \ref{fig:sky}, the depth prediction of textureless region around the foreground object is polluted, which results in ambiguous boundary of the foreground object. This blooming effect is driven by the edge-aware smoothness loss \cite{ranjan2019competitive} and appears more likely on objects with intricate shape.
\\\\
\noindent \textbf{Effect of Pretraining}
As shown in Table \ref{tab:ablation_ATE}, through the comparison of all the training strategies with/without weights pretrained on KITTI, we find out using pretrained model on other dataset does not explicitly improves the SLAM performance. This reveals that the transferability of Monodepth2 (with ResNet-18 as depth encoder) is limited. However, pretrained model guarantees the stability and robustness of RGB-D based tracking, while tracking failure continues to happen on all the sequence using the model specifically from monocular training (M) without pretrained on KITTI. Obviously, the depth and scale ambiguity is not learned by monocular training (M) standalone.

As stated above, we recommend interested readers to utilize Monodepth2 with the mixed training strategy and pretrained weights (MS*) to reproduce our work and research on similar agriculture scenes.

\subsection{Dense Reconstruction}
\begin{figure}[!t]
  \centering
  \includegraphics[width=0.95\linewidth]{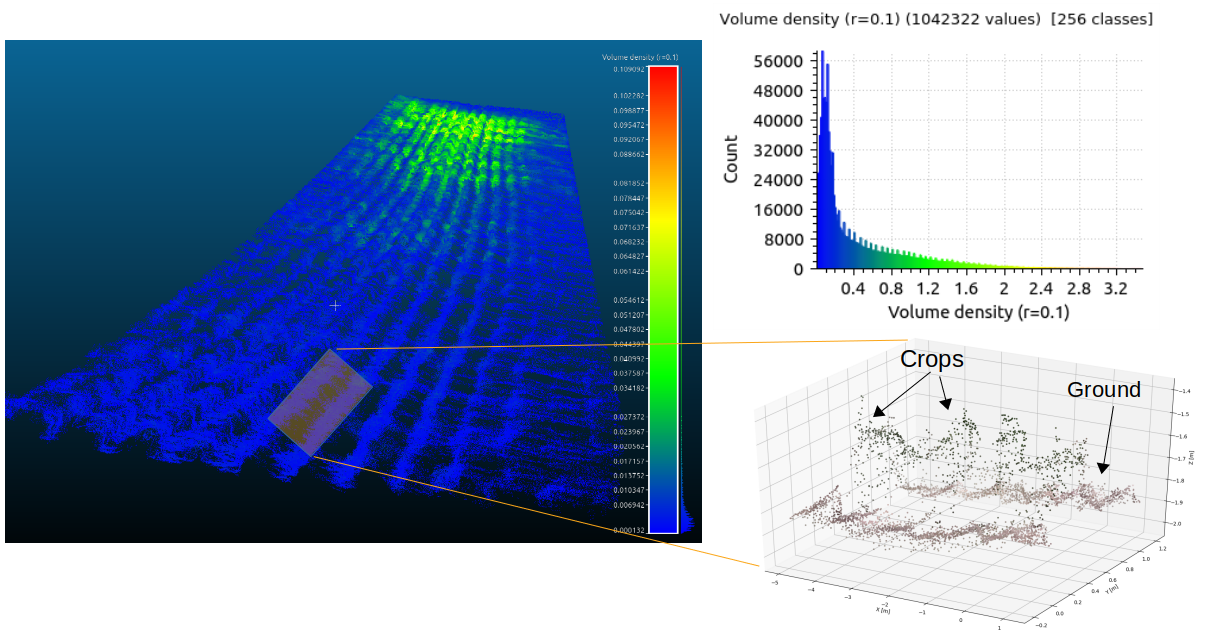}
  \caption{\textbf{Volume density ($R=0.1$ m) of the dense point cloud shown in Figure \ref{fig:colmap}.} Left: the density heatmap of the point cloud; Top-right: the histogram of volume density; Bottom-right: a subset of the dense point cloud.}
  \label{fig:geometric_result}
\end{figure}

In general, MVS can be initiated either with SfM or visual SLAM depending on whether the input data is an ordered sequence or unordered images, which means one of the pre-conditions is the poses of the images can be successfully recovered beforehand. In this work, we are able to reconstruct the dense point cloud offline (Figure \ref{fig:colmap}) up to real scale after geo-registration, where the potential drifts are eliminated by GPS measurement. However, the employed real-time algorithm REMODE estimates depth based on depth filter, which approximates the mean and variance of the depth at each pixel's position and updates the depth uncertainty when there is a new measurement (new image captured from the camera). The implementation of depth filter naturally requires a high frame rate to converge the depth uncertainty which is not the case regarding dataset Rosario (15Hz). While we are still able to reconstruct coarse dense point cloud on the fly using REMODE (Figure \ref{fig:remode}), a potential improvement could be to initialize the depth filter according to the depth estimated from CNN (e.g. consider depth estimated from Monodepth2 as the prior knowledge of the scene geometry) to accelerate convergence, as discussed in \cite{loo2019cnn}.
\\\\
\noindent \textbf{Point Cloud and Density}
The volume density is calculated using CloudCompare \cite{CloudCompare}, which is an open-source 3D point cloud and mesh processing software (see Figure \ref{fig:geometric_result}). The density of the map stays relatively constant throughout the sequence, except during slowdowns and stops. In those cases, more keyframes are taken within the same area, increasing the density of the map in this region. Moreover, we illustrate on a very small subset of the dense point cloud, where we can see the height of the crops. The crops and ground can be easily recognized, separated, and measured, which provides very valuable information.

\section{Conclusion}
\label{sec:conclusion}

Our work successfully presented a monocular vision-based architecture for mapping and localization explored under challenging agricultural environment, with new baselines provided for the relevant research community. 
Future works can explore other types of indirect SLAM systems, such as ones integrating GPS, or IMU, thus leveraging the advantages of feature-based tracking described here without the drifting issue. 

\section{Acknowledgment}

The research leading to these results has been partially funded by the German BMBF project MOVEON (Funding reference number 01IS20077) and by the German BMBF project SocialWear (Funding reference number 01IW20002).

\clearpage
{\small
\bibliographystyle{ieee_fullname}
\bibliography{bib}
}

\end{document}